\documentclass[10pt]{article}
\usepackage{spconf,amsmath,graphicx,multirow}
\usepackage[colorlinks,linkcolor=red]{hyperref}
\usepackage{booktabs}
\usepackage{anyfontsize}
\usepackage{amssymb}
\usepackage{enumitem}
\usepackage{caption}
\usepackage{hyperref}

\ninept
\title{Pixel-Superpixel Contrastive Learning and Pseudo-Label Correction for Hyperspectral Image Clustering}
\name{Renxiang Guan$^{\dag}$$^{1}$, Zihao Li$^{\dag}$$^{2}$, Xianju Li$^{*2}$, Chang Tang$^{2}$
\thanks{$^{\dag}$ Equally contributed.\quad $^{*}$ Corresponding author.}
\thanks{
This work was supported in part by the National Natural Science Foundation of China under Grants No. U21A2013 and 42071430. 
}}
\address{
$^1$College of Computer, National University of Defense Technology, Changsha 410073, China;\\
$^2$School of Computer Science, China University of Geosciences, Wuhan, 430074, China \\
renxiangguan@nudt.edu.cn; ddwhlxj@cug.edu.cn
}

\begin{document}

%
\maketitle

\begin{abstract}

Hyperspectral image (HSI) clustering is gaining considerable attention owing to recent methods that overcome the inefficiency and misleading results from the absence of supervised information. Contrastive learning methods excel at existing pixel-level and superpixel-level HSI clustering tasks. The pixel-level contrastive learning method can effectively improve the ability of the model to capture fine features of HSI but requires a large time overhead. The superpixel-level contrastive learning method utilizes the homogeneity of HSI and reduces computing resources; however, it yields rough classification results. To exploit the strengths of both methods, we present a pixel–superpixel contrastive learning and pseudo-label correction (PSCPC) method for the HSI clustering. PSCPC can reasonably capture domain-specific and fine-grained features through superpixels and the comparative learning of a small number of pixels within the superpixels. To improve the clustering performance of superpixels, this paper proposes a pseudo-label correction module that aligns the clustering pseudo-labels of pixels and superpixels. In addition, pixel-level clustering results are used to supervise superpixel-level clustering, improving the generalization ability of the model. Extensive experiments demonstrate the effectiveness and efficiency of PSCPC.
\end{abstract}

\section{Introduction}
\label{sec:introduction}

With the rapid development of hardware imaging technologies, hyperspectral remote sensing technology utilizes nanoscale imaging spectrometers for simultaneous imaging of ground objects in dozens or hundreds of bands \cite{2019HSI_Class_Review, 10335719}. Hyperspectral imagery (HSI) acquired by an imaging spectrometer contains rich spatial, radiative, and spectral information of ground objects, realizing the nature of graph unification \cite{2023SAPC, 10123082}. 

In the field of HSI analysis and processing, the classification of ground objects using remote-sensing data is an important research direction \cite{2022ResCapsNet, 2023AMF-GCN}. Based on whether it contains label information, HSI classification includes supervised and unsupervised classification. In practical applications, it is difficult to obtain a large number of labeled training samples; therefore, unsupervised classification or clustering methods have a wide range of applications. HSI clustering divides unlabeled hyperspectral pixels into multiple nonoverlapping clusters such that the pixels within a cluster are as similar as possible, whereas the pixels between clusters are as different as possible \cite{2021HSI_Cluster_Review}.

Traditional clustering algorithms have been widely used in HSI clustering tasks, such as $k$-means \cite{1979kmeans}, fuzzy $c$-means (FCM) \cite{2013FCM}, and subspace clustering \cite{ou2020anchor, zhou2023}. However, when these classical algorithms cluster high-dimensional data, such as HSI, it is difficult to extract discriminable visual feature representations, and satisfactory clustering results may not be obtained \cite{2023DSCRLE}. To solve this problem, many researchers have proposed the use of deep neural networks to learn the feature representation of HSI, combined with traditional clustering algorithms, to achieve deep clustering of image data \cite{2022DGAE, 2023SAPC}. The deep clustering algorithm optimizes the feature space and clustering results simultaneously, makes full use of the rich spectral and spatial information in the HSI, achieving good results. The main problem to be solved using a deep clustering algorithm is learning a discriminative representation that produces favorable results \cite{2021GRRSC}.

Fortunately, emerging contrastive learning, which can learn representative features that are more essential and higher dimensional, has received considerable attention in the unsupervised tasks of natural images \cite{2020MOCO, 2022TMI} and HSI \cite{2022NCSC, 2023CMSCGC}. The main idea of contrastive learning is to compare the data with semantically similar examples (positive examples) and semantically dissimilar examples (negative examples). By designing the model structure and comparison loss, similar examples in the feature space are closer and dissimilar; the instances of the distribution are far away, so as to better learn the characteristics of the data \cite{2020SimCLR}. Existing hyperspectral contrastive learning algorithms can be classified into pixel- and superpixel-level contrasts \cite{2021SSCC, 2022NCSC} at the input data level. Pixel-level comparison algorithms can achieve finer land classification but require more computing resources. On the other hand, superpixel-level contrastive learning algorithms reduce the amount of calculation but affect the clustering accuracy to a certain extent and also loses the spatial smoothness of features.

To exploit the strengths of both worlds based on superpixel contrastive learning, a small number of pixels within the superpixels are randomly selected for pixel-level contrastive learning. The advantage lies in bringing pixel representations within the same superpixel closer together and representations within different superpixels farther away, resulting in smoother clustering. In addition, to further improve the accuracy of superpixel-level clustering, a false-label correction mechanism is proposed that uses the clustering results of pixels to generate pseudo-labels of superpixels to promote local and global consistency of the clustering results and achieve more refined clustering. Extensive experiments have demonstrated that the proposed algorithm outperforms other related algorithms.

\section{contrastive learning}
\label{sec:Review}

Objects are believed to have certain general features. Contrastive learning learns such features, which have strong transferability, and can achieve good results in the clustering of unlabeled data. The core idea of contrastive learning algorithms is to use sample information to generate a supervisory signal and extract better modal features by continuously shortening the distance between positive samples and pushing away the distance between negative samples. After constructing the positive and negative samples and extracting the features, it is necessary to set a loss function to reduce the distance between the positive and negative example pairs in the representation space. The loss function generally uses the InfoNCE loss \cite{2018InfoNCE}, and the loss function definition formula for the positive example pair $(z_{i}, z_{j})$ is

\begin{equation}
l=-\log \frac{\exp \left(\frac{\operatorname{sim}\left(z_{i}, z_{j}\right)}{\tau}\right)}{\left.\exp \left(\frac{\operatorname{sim}\left(z_{i}, z_{j}\right)}{\tau}\right)+\sum_{k=1}^K \exp \left(\frac{\operatorname{sim}\left(z_{i}, z_{k}\right)}{\tau}\right)\right)},
\end{equation}
where $sim(z_{i}, z_{j})$ is a similarity function that calculates the similarity between $z_{i}$ and the positive sample $z_{j}$, $z_{k}$, $k \in[1, K]$ is the matching K negative samples, and $\tau$ is the temperature coefficient.

\section{Proposed approach}
\label{sec:approach}

\begin{figure*}[htb]
  \centering
  \centerline{\includegraphics[width=15cm]{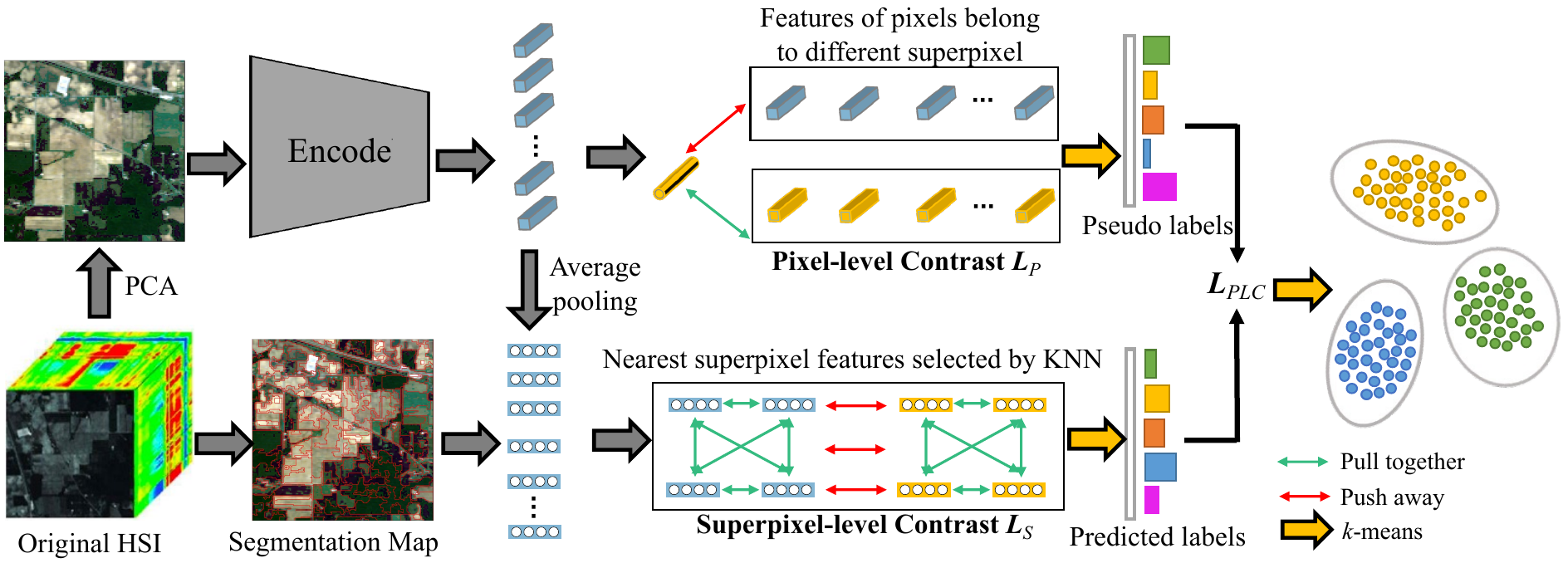}}
\caption{Illustration of our proposed contrastive learning framework.}
\label{fig:main}
\end{figure*}

The PSCPC model proposed in the paper is shown in \ref{fig:main}. After the input HSI is preprocessed and encoded, pixel-level contrast, superpixel level contrast and soft label correction modules are used to extract robust feature. Each module will be introduced separately.

\subsection{System pipeline}

As our primary goal is not the design of a specific model, we employ a commonly used encoder–decoder architecture. The input image patch is compressed into a low-dimensional representation by the encoder, which is then decoded back to its original dimensions by a decoder. The encoder used was a ResNet-18 \cite{2016ResNet} backbone. Before inputing into the model, we used principal component analysis \cite{2005pca} to perform band dimensionality reduction on the HSI to reduce redundant information interference.

\subsection{Superpixel-Level Contrastive Learning}

Prior knowledge of HSI research shows that hyperspectral data contain rich local spatial structure information and certain homogeneity \cite{2022NCSC}, and performing tasks at the superpixel level can help save resources and better aggregate spatial information.

Superpixels are the result of image segmentation; each superpixel is a group of pixels with similar properties. The estimation of scale parameter (ESP) method \cite{2010ESP} was used for the image segmentation of remotely sensed data. The HSI is segmented into superpixels to obtain superpixels $S=\left\{s_1, s_2, \ldots, s_N\right\}$, where N is the number of superpixels.

The features of a superpixel are the mean of the feature maps of the pixels extracted by the encoder. The feature-averaged pooling steps are as follows:

\begin{equation}
\stackrel{}{H}_n=\frac{1}{\left|\Omega_{s_i}\right|} \sum_{k \in \Omega_{s_i}}{\stackrel{}{h_k}},
\end{equation}
where $\Omega_{s_i}$ represents the set of pixels of the superpixel, and $h$ represents the pixel feature vector. Otherwise, we combined the deep features extracted by the encoder with the shallow attribute features of the superpixels. Specifically, we fuse the pooled feature map with the geometric features (area, perimeter, and aspect ratio) of the superpixel and its center point. We normalize these features according to the category. 

Different from the standard comparative learning method, the standard comparative learning method usually uses data between multiple views for comparative learning. In this study, the $k$-nearest neighbor algorithm is used to capture the nearest $k$ samples as positive samples, because samples with closer feature distances are also more similar. Thus, each sample has $k$ positive and $N-k-1$-negative samples. The goal of superpixel contrastive loss is to maximize the distance between $N-k-1$ positive samples, and its loss is
\vspace{-0.2cm}
\begin{equation}
\mathcal{L}_{\mathrm{S}}=\frac{1}{N} \sum_{i=1}^N \sum_{j \in \mathcal{N}_k(i)}-\log \frac{\exp \left(\operatorname{sim}\left({{H}}_i, {{H}}_{j}\right) / \tau\right)}{\sum_{t=1}^N \left[\mathrm{1}_{t \neq i} \exp \left(\operatorname{sim}\left({{H}}_i, {{H}}_{t}\right) / \tau\right)\right]},
\end{equation}
where sim(·) and $\tau$ are defined as in Eq. (1), and $\mathrm{}{1}_{j \neq i}=\{0,1\}$ is an indicator function.

In this manner, the learned representation is discriminative with the help of contrastive learning and the properties of the superpixels. Compared with the construction of multiview positive and negative sample comparison methods \cite{2020SimCLR}, our method reduces the data augmentation process. In addition, our positive samples not only consider a pair, but also consider the adjacent $k$ samples, which will clearly attract similar samples and repel dissimilar samples, and generate robust feature representations.

\subsection{Pixel-Level Contrastive Learning}

The essence of HSI clustering is to divide similar pixels into closer clusters. Thus, improving pixel-level representation learning can improve the clustering accuracy because it encourages spatially close pixels to have similar representations. To obtain smooth features and fine classifications, we propose a pixel-level contrastive learning method. Because most of the pixels in a superpixel belong to the same class, to save computing resources, we randomly select $m$ pixels in each superpixel for comparison with the superpixel feature $H$. The loss function is expressed as follows:
\vspace{-0.2cm}
\begin{center}
\begin{equation}
l\left({s}_n\right)=-\log \frac{\sum_{t \in \Omega_{{s}_n}} \exp \left(\operatorname{sim}\left({{H}}_n, {{h}}_t\right) / \tau\right)}{\sum_{i \in {\Omega}} \exp \left(\operatorname{sim}\left({{H}}_n, {{h}}_i\right) / \tau\right)},
\end{equation}
\end{center}
where $H$ and $h$ are the superpixel and pixel features in Eq. (2), respectively. In the pixel-level comparative experiments, the positive samples were the remaining pixels in the same superpixel and the negative samples were the pixels that were not in the same superpixel. By minimizing Eq. (4), the consistency of the features within a superpixel is guaranteed while maintaining contrast with the rest of the region. The overall pixel-level contrastive loss is

\begin{equation}
\mathcal{L}_{\mathrm{P}}=\frac{1}{N} \sum_{i=1}^N l\left({s}_i\right)
\end{equation}
Pixel-level contrastive learning helps to group neighboring relevant components into the same region, which is critical for HSI clustering. Therefore, the overall contrastive loss is the sum of the superpixel- and pixel-level contrastive losses.

\begin{equation}
\mathcal{L}_{\mathrm{C}}=\mathcal{L}_{\mathrm{S}}+\mathcal{L}_{\mathrm{P}}
\end{equation}

\subsection{Correction Module for Pseudo-labels}

Compared to pixel-level clustering, the fineness of clustering is lower owing to the large input scale of superpixel-level tasks. In addition, the superpixel-level contrastive loss considers all but the closest $k$ samples as negative samples, which may separate similar samples and destroy the clustering structure. To improve the robustness of the clustering model, we propose a pseudo-label correction module.

Specifically, each superpixel has two labels. First, the $k$-means clustering algorithm is directly used to obtain predicted-label for each superpixel. Second, for the $M$ pixels in the superpixel, we calculate pseudo-label $\hat{\mathbf{y}}$ of the superpixel according to the clustering result, that is, the proportion of each type of pixel. To avoid the prediction results being too absolute, the prediction labels and pseudo labels here are both soft labels. The purpose of the pseudo-label correction module is to ensure that the results obtained by the two methods are consistent. The loss function of the pseudo-label correction module is

\begin{equation}
\mathcal{L}_{PLC}=\frac{1}{N} \sum_{n \in N} E\left(\hat{\mathbf{y}}, f_\theta(H)\right)
\end{equation}
where $E(p, q)$ denotes the cross-entropy between $p$ and $q$, and $f_\theta(H)$ denotes the $k$-means algorithm that clusters superpixel features to obtain superpixel soft labels. In this method, if the cluster label of the superpixel is consistent with that of the pixel in the superpixel, it is determined that the superpixel sample is clean; otherwise, it is unclean. The $\mathcal{L}_{PLC}$ loss is used to reduce the number of unclean samples so that the prediction results of the superpixels and pixels are aligned and the robustness of the model is improved.

Finally, the overall loss of the algorithm is $\mathcal{L}$:
\begin{equation}
\mathcal{L}=\mathcal{L}_{C} + \lambda\mathcal{L}_{PLC},
\end{equation}
where the parameter $\lambda$ balances the weights between the contrastive loss and pseudo-labels loss, and the $\lambda$ will be optimized in the experiments.

\begin{table*}[]
\caption{Clustering results of various methods on three benchmark datasets. The best results are shown in bold.}
\label{tab:results}
\begin{tabular}{c|c|cc|ccccc|ccc}
\hline
\multirow{2}{*}{\textbf{Datasets}} & \multirow{2}{*}{\textbf{Metrics}} & \multirow{2}{*}{\textbf{$k$-means}} & \multirow{2}{*}{\textbf{FCM}} & \multirow{2}{*}{\textbf{AEC}} & \multirow{2}{*}{\textbf{MS$^2$A-Net}} & \multirow{2}{*}{\textbf{NCSC}} & \multirow{2}{*}{\textbf{SSCC}} & \multirow{2}{*}{\textbf{DGAE}} & \multicolumn{3}{c}{\textbf{Ablation}} \\ \cline{10-12} 
 &  &  &  &  &  &  &  &  & \textbf{w/o $\mathcal{L}_{C}$} & \textbf{w/o $\mathcal{L}_{PLC}$} & \textbf{PSCPC}  \\
 \hline
\multirow{3}{*}{\textbf{Indian}} & \textbf{ACC} & 0.4016 & 0.3485 & 0.3458 & 0.4197 & 0.4829 & 0.5286 & 0.6002 & 0.5113 & 0.6344 & \textbf{0.6446} \\
 & \textbf{NMI} & 0.4480 & 0.4102 & 0.4172 & 0.4111 & 0.4392 & 0.5521 & 0.6433 & 0.5637 & 0.6426 & \textbf{0.6452} \\
 & \textbf{Kappa} & 0.3691 & 0.2905 & 0.2775 & 0.3027 & 0.5177 & 0.4793 & \textbf{0.6129} & 0.4579 & 0.5927 & 0.6032 \\
 \hline
\multirow{3}{*}{\textbf{PaviaU}} & \textbf{ACC} & 0.5261 & 0.5541 & 0.6281 & 0.6289 & 0.4392 & 0.4931 & 0.6221 & 0.5779 & 0.6906 & \textbf{0.7316} \\
 & \textbf{NMI} & 0.5334 & 0.5815 & 0.5808 & 0.5653 & 0.2952 & 0.3745 & 0.5718 & 0.4382 & 0.5745 & \textbf{0.6194} \\
 & \textbf{Kappa} & 0.4337 & 0.4116 & 0.6068 & 0.4960 & 0.4355 & 0.5290 & 0.5822 & 0.3497 & 0.5454 & \textbf{0.6366} \\
 \hline
\multirow{3}{*}{\textbf{SalinesA}} & \textbf{ACC} & 0.8121 & 0.8044 & 0.8292 & 0.8017 & 0.7332 & 0.9036 & 0.8604 & 0.8338 & 0.8622 & \textbf{0.9062} \\
 & \textbf{NMI} & 0.8125 & 0.8019 & 0.8557 & 0.6733 & 0.7028 & 0.9023 & 0.8621 & 0.8703 & 0.9046 & \textbf{0.9169} \\
 & \textbf{Kappa} & 0.7702 & 0.7822 & 0.7526 & 0.7222 & 0.6572 & \textbf{0.8393} & 0.8353 & 0.7929 & 0.8177 & 0.8305 \\
 \hline
\end{tabular}
\end{table*}

\section{Experiments}
\label{sec:experiment}
\subsection{Experimental Setup}

\textbf{Datasets.} Three sets of real HSI were used in the experiments: Indian Pines, Pavia University, and Salinas-A datasets. In particular, the Indian Pines and Salinas-A datasets were collected using the airborne visible/infrared imaging spectrometer (AVIRIS) sensor. The Indian Pines dataset contains 220 bands with a spatial resolution of 20m and an image size of 145×145. The Salinas-A dataset is a subscene of the Salinas image with a size of 86×86 pixels, a spatial resolution of 3.7 m, and 224 bands. The Pavia University dataset was collected using the reflective optics system imaging spectrometer (ROSIS) sensor and contains 115 bands with an image size of 610×340 pixels.

\textbf{Compared Methods.} To test the performance of the proposed algorithm, we compare several traditional and deep clustering algorithms with the proposed PSCPC. They are $k$-means \cite{1979kmeans}, FCM \cite{2013FCM}, autoencoder clusering (AEC) \cite{1993autoencoder}, multiscale spectral-spatial association network (MS2A-Net) \cite{2022S2ANet}, spectral-spatial contrastive clustering (SSCC) \cite{2021SSCC}, dual Graph autoencoder (DGAE) \cite{2022DGAE}, and neighborhood contrastive subspace clustering (NCSC) \cite{2022NCSC}.

\textbf{Evaluation Metrics.} For the experimental results of the different algorithms, three standard metrics were used to evaluate the clustering performance: overall accuracy (OA), Kappa coefficient, and normalized mutual information (NMI). Their ranges were in the interval [0,1], and the higher the score, the better the clustering performance. All reported results were obtained by averaging 10 trials.

\begin{figure}[htb]
\centering
\centerline{\includegraphics[width=8.5cm]{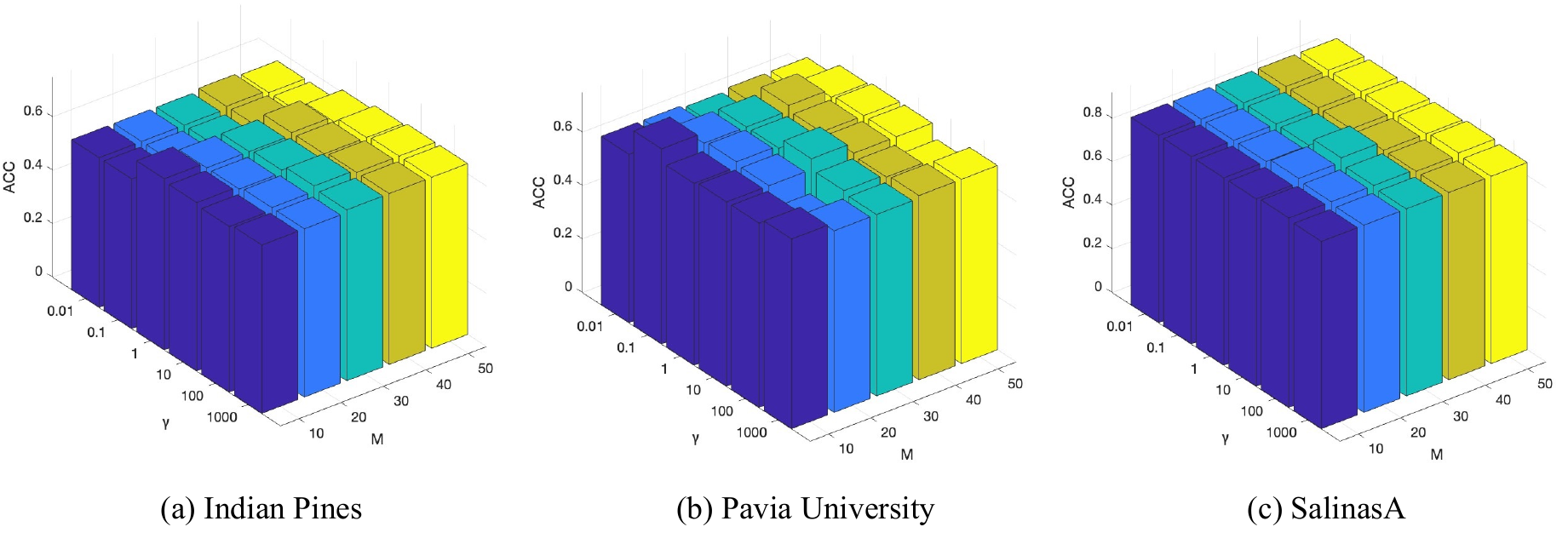}}
\caption{Model sensitivity with the variation of two hyper-parameters. The X-axis, Y-axis, and Z-axis refer to the M value, $\lambda$ value, and accuracy performance, respectively.}
\label{fig:param}
\vspace{-0.5 cm}
\end{figure}

\textbf{Parameter Settings.} There are two important parameters in pixel–superpixel contrastive learning and pseudo-label correction (PSCPC): the parameter $\lambda$ and the number $M$ of randomly selected pixels in each superpixel, where parameter $\lambda$ balances the weights between the contrastive loss and pseudo-labels. Figure \ref{fig:param} shows the variation trend of the clustering accuracy OA for the parameters $\lambda$ and $M$. The value range of the parameter $\lambda$ was set to [1e-2, 1e-1, 1, 1e2, 1e3], and the value range of the parameter $M$ ranged from 10 to 50 in increments of 10. As shown in Figure \ref{fig:param}, the highest accuracy was obtained when they were selected as 10, 10, and 100 for parameter $\lambda$. For parameter M, the highest accuracy was obtained when the three datasets were selected as 30, 10, and 30.

\begin{figure}[htb]
\centering
\centerline{\includegraphics[width=8.5cm]{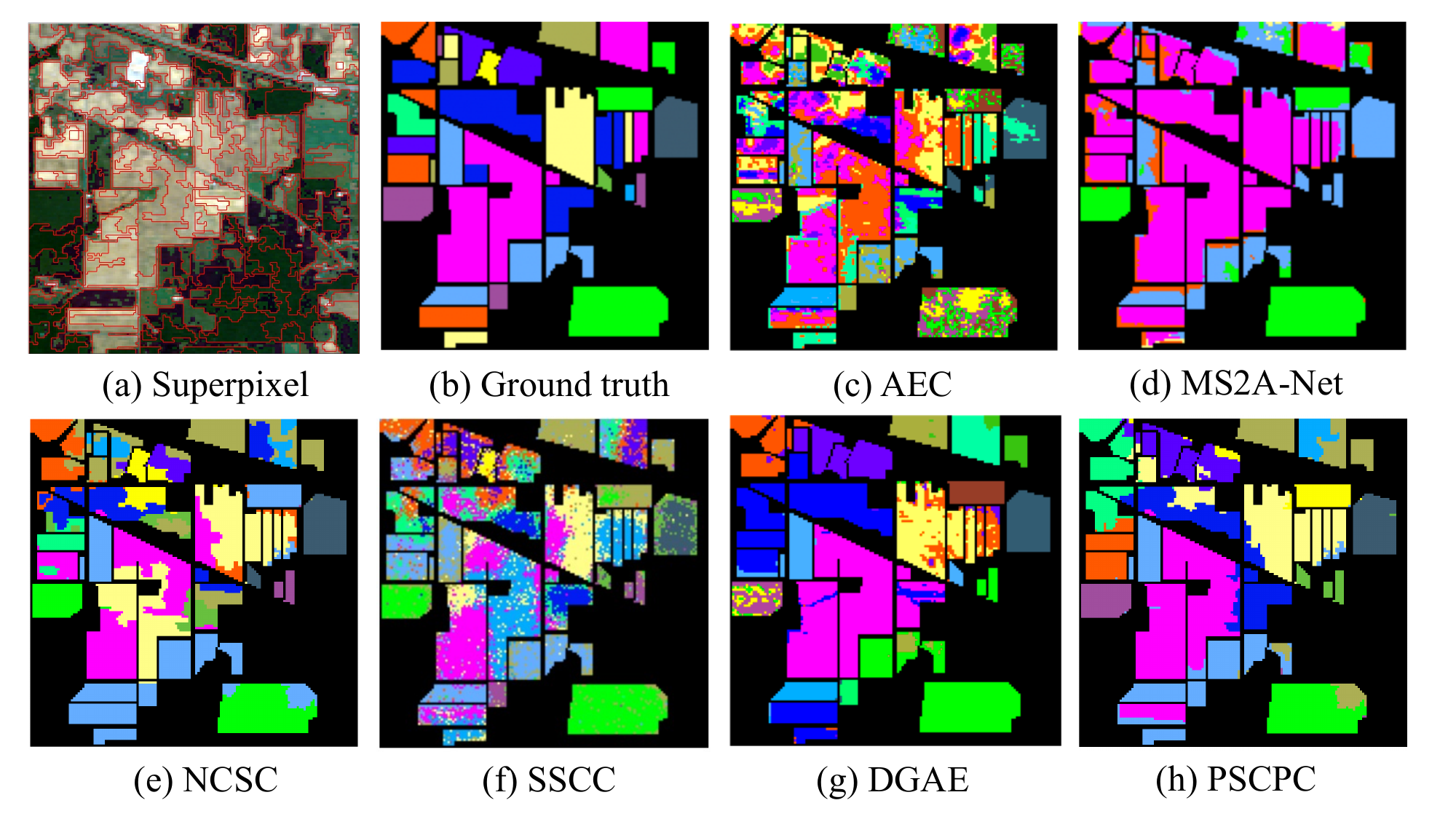}}
\caption{Clustering results of different deep clustering algorithms on the Indian Pines dataset.}
\label{fig:visal}
\vspace{-0.5 cm}
\end{figure}

\subsection{Experimental results}
\label{sec:experiment}
\textbf{Clustering Results.}
In this section, we describe the extensive experiments conducted to compare the proposed PSCPC with classical and deep methods. Table \ref{tab:results} reports the clustering results for OA, NMI, and Kappa obtained using different clustering methods on the three HSI datasets, where we ran all methods 10 times and reported their average results. It can be observed from the table that the proposed PSCPC method achieved the best clustering performance on all test datasets. For example, on the Indian Pines dataset, the performance improvements over the second-best method, DGAE, were 4.44$\%$ and 0.19$\%$ for ACC and NMI, respectively. On the Pavia University dataset, PSCPC achieved 14.27$\%$, 5.41$\%$, and 14.06$\%$ improvements in OA, NMI, and Kappa, respectively, significantly outperforming the second-best MS$^2$A-Net. To observe the clustering performance of the model more intuitively, this study presents a visual classification diagram of the PSCPC and five deep clustering methods. As shown in Figure \ref{fig:visal}, the classification map obtained by PSCPC has fewer misclassified points and the distribution of ground objects is smoother, which proves the efficiency and feasibility of PSCPC in HSI clustering.

\begin{table}[]
\setlength{\tabcolsep}{3pt}
\caption{Running time (second) on three HSI datasets.}
\begin{tabular}{c|cccccc}
\hline
\textbf{Datasets} & \textbf{AEC} & \textbf{MS$^2$A-Net} & \textbf{NCSC} & \textbf{SSCC} & \textbf{DGAE} & \textbf{PSCPC} \\
\hline
\textbf{Indian}   & 121.8       & 128.5            & 1825        & 3822        & 158.4        & 240.8         \\

\textbf{PaviaU}   & 298.2       & 416.5            & 6813        & 5147        & 293.7        & 292.1         \\

\textbf{SalinesA} & 63.28       & 38.14            & 519.9        & 1700        & 48.34        & 52.36        \\
\hline
\end{tabular}
\vspace{-0.5cm}
\label{tab:time}
\end{table}

\textbf{Ablation Study.} To further verify the impact of the contrastive learning and pseudo-label correction modules, the contrastive learning and pseudo-label correction modules were removed and then tested on the three datasets. The results are shown in Table \ref{tab:results}, where w/o $\mathcal{L}_{C}$ is the clustering result when only the comparative learning module is removed, and w/o $\mathcal{L}_{PLC}$ is the clustering result when only the pseudo-label correction module is removed. The ablation experiment results in Table \ref{tab:results} show that the clustering results obtained by retaining only the pseudo-label correction module or contrastive learning module were worse than those obtained by PSCPC. The results show that the comparative learning module can effectively improve the ability of the model to acquire features and confirm the correction effect of the pseudo-label correction module; the combination of the two can effectively improve the clustering performance. 

\textbf{Time Cost.} In this section, we record the average running time (in seconds) consumed by all deep clustering methods. As displayed in Table \ref{tab:time}, on the Pavia University dataset, PSCPC achieved the best results, and other datasets also showed competition. In contrast, because NCSC and SSCC are subspace-based clustering methods, it is difficult to compare them with $k$-means-based methods in terms of running speed.

\textbf{Parameter Analysis.} To verify the influence of different parameters on the PSCPC, we conducted parameter sensitivity studies on the three HSI datasets. As can be seen from Figure \ref{fig:param}, on the Indian Pines dataset, when $\gamma$ is greater than one, the fluctuation in the model performance is small. In addition, different datasets have different requirements for optimal $M$, which is related to the number of superpixels segmented from the HSI datasets. In summary, the model proposed in this study is insensitive to these two parameters.

\section{Conclusion}
In this paper, a novel HSI clustering method named PSCPC was proposed. Specifically, we conducted contrastive learning at the superpixel and a few pixel levels to encourage the model to learn homogeneous and smooth features. In addition, to improve the accuracy of superpixel-level clustering, we proposed a pseudo-label correction module that brings the labels formed by pixel clustering in superpixels and the results of superpixel clustering closer together, thereby achieving better clustering performance. The clustering tasks, ablation studies, parameter sensitivity analysis, experimental time analysis, and cluster visualization analysis performed on the three datasets in this study demonstrate the effectiveness and superiority of PSCPC. In future research, we will consider introducing the anchor concept into the superpixel clustering task to further shorten the clustering time. Simultaneously, multiview information of the HSI can be considered to learn more discriminative features.
\label{sec:conclusions}

\clearpage
\bibliographystyle{IEEEbib}
\bibliography{strings,refs}

\begin{thebibliography}{10}

\bibitem{2019HSI_Class_Review}
Shutao Li, Weiwei Song, Leyuan Fang, Yushi Chen, Pedram Ghamisi, and Jón~Atli
  Benediktsson,
\newblock ``Deep learning for hyperspectral image classification: An
  overview,''
\newblock {\em IEEE Transactions on Geoscience and Remote Sensing}, vol. 57,
  no. 9, pp. 6690--6709, 2019.

\bibitem{10335719}
Yingjie Tang, Shou Feng, Chunhui Zhao, Yuanze Fan, Qian Shi, Wei Li, and Ran
  Tao,
\newblock ``An object fine-grained change detection method based on frequency
  decoupling interaction for high-resolution remote sensing images,''
\newblock {\em IEEE Transactions on Geoscience and Remote Sensing}, vol. 62,
  pp. 1--13, 2024.

\bibitem{2023SAPC}
Guozhu Jiang, Jie Zhang, Yongshan Zhang, Xinwei Jiang, and Zhihua Cai,
\newblock ``Structured-anchor projected clustering for hyperspectral images,''
\newblock in {\em ICASSP 2023-2023 IEEE International Conference on Acoustics,
  Speech and Signal Processing (ICASSP)}. IEEE, 2023, pp. 1--5.

\bibitem{10123082}
Chunhui Zhao, Yingjie Tang, Shou Feng, Yuanze Fan, Wei Li, Ran Tao, and Lifu
  Zhang,
\newblock ``High-resolution remote sensing bitemporal image change detection
  based on feature interaction and multitask learning,''
\newblock {\em IEEE Transactions on Geoscience and Remote Sensing}, vol. 61,
  pp. 1--14, 2023.

\bibitem{2022ResCapsNet}
Renxiang Guan, Zihao Li, Teng Li, Xianju Li, Jinzhong Yang, and Weitao Chen,
\newblock ``Classification of heterogeneous mining areas based on rescapsnet
  and gaofen-5 imagery,''
\newblock {\em Remote Sensing}, vol. 14, no. 13, pp. 3216, 2022.

\bibitem{2023AMF-GCN}
Jie Liu, Renxiang Guan, Zihao Li, Jiaxuan Zhang, Yaowen Hu, and Xueyong Wang,
\newblock ``Adaptive multi-feature fusion graph convolutional network for
  hyperspectral image classification,''
\newblock {\em Remote Sensing}, vol. 15, no. 23, pp. 5483, 2023.

\bibitem{2021HSI_Cluster_Review}
Han Zhai, Hongyan Zhang, Pingxiang Li, and Liangpei Zhang,
\newblock ``Hyperspectral image clustering: Current achievements and future
  lines,''
\newblock {\em IEEE Geoscience and Remote Sensing Magazine}, vol. 9, no. 4, pp.
  35--67, 2021.

\bibitem{1979kmeans}
John~A Hartigan and Manchek~A Wong,
\newblock ``Algorithm as 136: A k-means clustering algorithm,''
\newblock {\em Journal of the royal statistical society. series c (applied
  statistics)}, vol. 28, no. 1, pp. 100--108, 1979.

\bibitem{2013FCM}
James~C Bezdek,
\newblock {\em Pattern recognition with fuzzy objective function algorithms},
\newblock Springer Science \& Business Media, 2013.

\bibitem{ou2020anchor}
Qiyuan Ou, Siwei Wang, Sihang Zhou, Miaomiao Li, Xifeng Guo, and En~Zhu,
\newblock ``Anchor-based multiview subspace clustering with diversity
  regularization,''
\newblock {\em IEEE MultiMedia}, vol. 27, no. 4, pp. 91--101, 2020.

\bibitem{zhou2023}
Sihang Zhou, Qiyuan Ou, Xinwang Liu, Siqi Wang, Luyan Liu, Siwei Wang, En~Zhu,
  Jianping Yin, and Xin Xu,
\newblock ``Multiple kernel clustering with compressed subspace alignment,''
\newblock {\em IEEE Transactions on Neural Networks and Learning Systems}, vol.
  34, no. 1, pp. 252--263, 2023.

\bibitem{2023DSCRLE}
Yang Zhao and Xuelong Li,
\newblock ``Deep spectral clustering with regularized linear embedding for
  hyperspectral image clustering,''
\newblock {\em IEEE Transactions on Geoscience and Remote Sensing}, vol. 61,
  pp. 1--11, 2023.

\bibitem{2022DGAE}
Yongshan Zhang, Yang Wang, Xiaohong Chen, Xinwei Jiang, and Yicong Zhou,
\newblock ``Spectral--spatial feature extraction with dual graph autoencoder
  for hyperspectral image clustering,''
\newblock {\em IEEE Transactions on Circuits and Systems for Video Technology},
  vol. 32, no. 12, pp. 8500--8511, 2022.

\bibitem{2021GRRSC}
Yaoming Cai, Meng Zeng, Zhihua Cai, Xiaobo Liu, and Zijia Zhang,
\newblock ``Graph regularized residual subspace clustering network for
  hyperspectral image clustering,''
\newblock {\em Information Sciences}, vol. 578, pp. 85--101, 2021.

\bibitem{2020MOCO}
Kaiming He, Haoqi Fan, Yuxin Wu, Saining Xie, and Ross Girshick,
\newblock ``Momentum contrast for unsupervised visual representation
  learning,''
\newblock in {\em Proceedings of the IEEE/CVF conference on computer vision and
  pattern recognition}, 2020, pp. 9729--9738.

\bibitem{2022TMI}
Zeyu Gao, Chang Jia, Yang Li, Xianli Zhang, Bangyang Hong, Jialun Wu, Tieliang
  Gong, Chunbao Wang, Deyu Meng, Yefeng Zheng, et~al.,
\newblock ``Unsupervised representation learning for tissue segmentation in
  histopathological images: From global to local contrast,''
\newblock {\em IEEE Transactions on Medical Imaging}, vol. 41, no. 12, pp.
  3611--3623, 2022.

\bibitem{2022NCSC}
Yaoming Cai, Zijia Zhang, Pedram Ghamisi, Yao Ding, Xiaobo Liu, Zhihua Cai, and
  Richard Gloaguen,
\newblock ``Superpixel contracted neighborhood contrastive subspace clustering
  network for hyperspectral images,''
\newblock {\em IEEE Transactions on Geoscience and Remote Sensing}, vol. 60,
  pp. 1--13, 2022.

\bibitem{2023CMSCGC}
Renxiang Guan, Zihao Li, Xianju Li, Chang Tang, and Ruyi Feng,
\newblock ``Contrastive multi-view subspace clustering of hyperspectral images
  based on graph convolutional networks,''
\newblock {\em arXiv preprint arXiv:2312.06068}, 2023.

\bibitem{2020SimCLR}
Ting Chen, Simon Kornblith, Mohammad Norouzi, and Geoffrey Hinton,
\newblock ``A simple framework for contrastive learning of visual
  representations,''
\newblock in {\em International conference on machine learning}, 2020, pp.
  1597--1607.

\bibitem{2021SSCC}
Yaoming Cai, Zijia Zhang, Yan Liu, Pedram Ghamisi, Kun Li, Xiaobo Liu, and
  Zhihua Cai,
\newblock ``Large-scale hyperspectral image clustering using contrastive
  learning,''
\newblock {\em arXiv preprint arXiv:2111.07945}, 2021.

\bibitem{2018InfoNCE}
Aaron van~den Oord, Yazhe Li, and Oriol Vinyals,
\newblock ``Representation learning with contrastive predictive coding,''
\newblock {\em arXiv preprint arXiv:1807.03748}, 2018.

\bibitem{2016ResNet}
Kaiming He, Xiangyu Zhang, Shaoqing Ren, and Jian Sun,
\newblock ``Deep residual learning for image recognition,''
\newblock in {\em Proceedings of the IEEE conference on computer vision and
  pattern recognition}, 2016, pp. 770--778.

\bibitem{2005pca}
Michael~D Farrell and Russell~M Mersereau,
\newblock ``On the impact of pca dimension reduction for hyperspectral
  detection of difficult targets,''
\newblock {\em IEEE Geoscience and Remote Sensing Letters}, vol. 2, no. 2, pp.
  192--195, 2005.

\bibitem{2010ESP}
Lucian Drǎgu{\c{t}}, Dirk Tiede, and Shaun~R Levick,
\newblock ``Esp: a tool to estimate scale parameter for multiresolution image
  segmentation of remotely sensed data,''
\newblock {\em International Journal of Geographical Information Science}, vol.
  24, no. 6, pp. 859--871, 2010.

\bibitem{1993autoencoder}
Geoffrey~E Hinton and Richard Zemel,
\newblock ``Autoencoders, minimum description length and helmholtz free
  energy,''
\newblock in {\em Advances in neural information processing systems}, 1993,
  vol.~6.

\bibitem{2022S2ANet}
Kasra~Rafiezadeh Shahi, Pedram Ghamisi, Behnood Rasti, Richard Gloaguen, and
  Paul Scheunders,
\newblock ``Ms2a-net: Multiscale spectral--spatial association network for
  hyperspectral image clustering,''
\newblock {\em IEEE Journal of Selected Topics in Applied Earth Observations
  and Remote Sensing}, vol. 15, pp. 6518--6530, 2022.

\end{thebibliography}

\end{document}